\documentclass{article}

     \usepackage[final]{neurips_2020}

\usepackage[utf8]{inputenc} 
\usepackage[T1]{fontenc}    
\usepackage{hyperref}       
\usepackage{url}            
\usepackage{booktabs}       
\usepackage{amsfonts}       
\usepackage{nicefrac}       
\usepackage{microtype}      
\usepackage[square, comma, sort&compress, numbers]{natbib}
\usepackage{graphicx}
\usepackage[ruled,vlined]{algorithm2e}
\title{Will Multi-modal Data Improves Few-shot Learning?}

\author{%
  Zilun Zhang\\
  \texttt{zilun.zhang@mail.utoronto.ca} \\
   \And
   Shihao Ma \\
   \texttt{rex.ma@mail.utoronto.ca} \\
   \And
   Yichun Zhang \\
   \texttt{yichun.zhang@mail.utoronto.ca} \\
}

\begin{document}

\maketitle

\begin{abstract}
  Most few-shot learning models utilize only one modality of data. We would like to investigate qualitatively and quantitatively how much will the model improve if we add an extra modality (i.e. text description of the image), and how it affects the learning procedure. To achieve this goal, we propose four types of fusion method to combine the image feature and text feature. To verify the effectiveness of improvement, we test the fusion methods with two classical few-shot learning models - ProtoNet and MAML, with image feature extractors such as ConvNet and ResNet12. The attention-based fusion method works best, which improves the classification accuracy by a large margin around 30\% comparing to the baseline result.
\end{abstract}

\section{Introduction}

In recent years, deep learning techniques have been applied and achieved great results in many domains such as computer vision and natural language processing. However, having large enough data to train on is essential for many deep learning applications. For some applications, it is often expensive or hard to collect enough training samples. Thus, few-shot learning research \cite{maml}\cite{protonet}  has gained increasing attention over recent years. Few-shot learning aims to make the model learn and generalize well with only a few training samples. However, most of the current few-shot learning models only utilize data from single modality, especially images. Even though there are a few works using multi-modal data for few-shot learning \cite{8451372} \cite{nortje2020direct}, they are mainly focusing on cross-modal generation tasks or image-text pair matching tasks. Whether adding an extra modality will improve the latent representation of the data, and further improve the model performance on downstream classification tasks remains under-explored.
In this work, we are trying to analyze this problem qualitatively and quantitatively, with the assumption of adding an extra modality (e.g. fine-grained text descriptions for corresponding images) will make the latent representation of each image class 

\begin{figure}[h!]
  \includegraphics[scale=0.25]{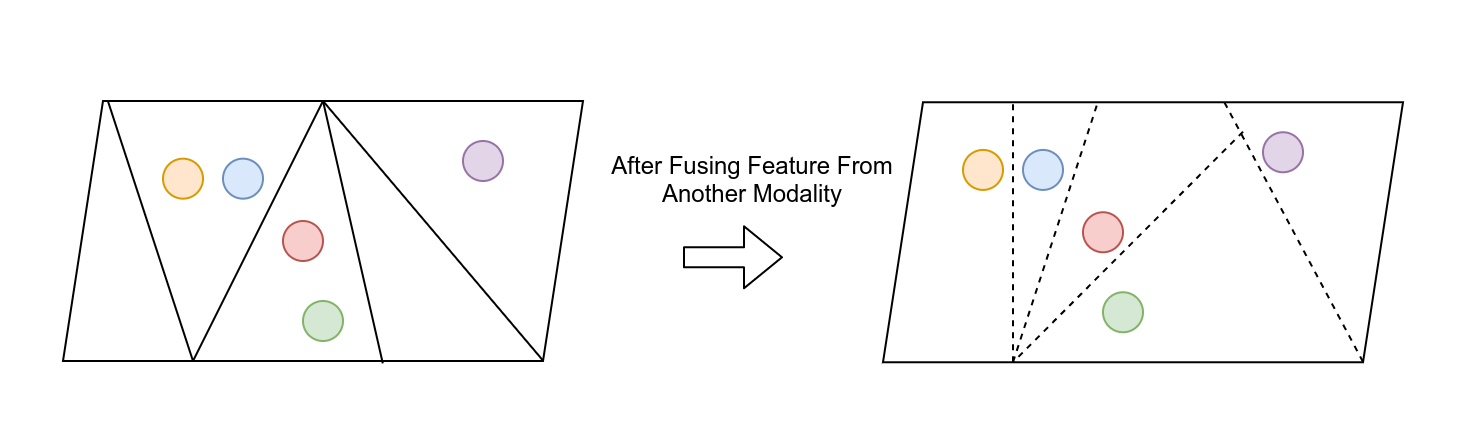}
  \caption{The idea of using multi-modal data for few-shot learning.}
  \label{fig:g1}
\end{figure}

 more discriminative therefore produce a better feature space and improve the few-shot classification accuracy as shown in Figure ~\ref{fig:g1}. Our contribution includes: \textbf{First}, qualitatively and quantitatively investigated how deep learning models improve with multi-modal data in the few-shot classification setup. \textbf{Second}, comparison of 4 different multi-modal fusion methods, including a novel attention-based fusing technique.




\section{Related Work}

\subsection{Few-shot Learning}
Many machine/deep learning models perform well when training with massive labeled data, but limited data could potentially reduce the power of machine/deep learning models. The core idea of few-shot learning is to create synthetic tasks to simulate the scenario in which only small amount of labeled data is given, and train the model with algorithms such as meta-learning to avoid over-fitting. MAML \citep*{maml} and Reptile \citep*{reptile} are gradient-based approaches that design the meta-learner as an optimizer that could learn to update the model parameters within few optimization steps given novel examples. Another type of method optimizes feature embeddings of input data using metric learning methods, such as ProtoNet \citep{protonet}, Relational Network \citep{relationnet}, Matching Network \citep{matchingnet} and DeepEMD \citep{deepemd}. Some works tackle the few-shot problem using graph-based methods. For each task, they make instances to be nodes, and relations between them to be edges. Then, graph-based methods such as GNN \citep{gnn}, TPN \citep*{liu2019learning} and DPGN \citep{dpgn} refine the node representations by aggregating and transforming neighboring nodes recursively. MetaOptNet \citep{metaopt} advocates the use of linear classifiers, which can be optimized as convex learning problems, instead of nearest-neighbor methods . LEO \citep{leo} utilizes an encoder-decoder architecture to mine the latent generative representations and predict high-dimensional parameters in extreme low-data regimes.

\subsection{Learning with Multi-Modal Data}
To better utilize the multi-modal data, many methods on how to fuse the multi-modal data are proposed. CentralNet \citep{centralnet} and SAL \citep{sal} fuse the multi-modal data using selective additive learning and weighted sum. Bilinear Pooling is adapted by \citep{bilinear} and \citep{bilinearpool}. They calculate the outer product of two modalities' representation and linearize it to a single vector. Some closely related areas such as VQA \citep{vqabaseline} and medical diagnosis \citep{Perrin2009} also have many works on fusing the multi-modal representation. RNN-based encoder-decoder models are used to assign weights to image features for the image caption task in \citep{sat}. This work \citep{wheretolook} divides each image into many different regions and calculates the attention between text embedding and the embedding of image regions. Stacked Attention Network \citep{stackedatt} performs a multi-step reasoning after the attention is calculated.

\section{Method}
\subsection{Problem Definition}
The goal of few-shot learning tasks with multi-model data is to train a model that can perform well in the case where only few multi-modal samples are given. The minimum unit during the training procedure is the episode, and each few-shot episode (task) has a \emph{support set} $\mathcal{S}$ and a \emph{query set} $\mathcal{Q}$. 
\\
Given training data $\mathbb{D}^{train}$, the support set $\mathcal{S} \subset \mathbb{D}^{train}$ contains $N$ classes with $K$ samples for each class (i.e., the $N$-way $K$-shot setting), it can be denoted as 
\begin{equation}
    \mathcal{S} = \{(x_{_1}^{image},\ x_{_1}^{text},\ y_{_1}),\ (x_{_2}^{image},\ x_{_2}^{text},\ y_{_2}),\ \dots,\ (x_{{_N} {_\times} {_K}}^{image},\ x_{{_N} {_\times} {_K}}^{text},\ y_{{_N} {_\times} {_K}})\}
\end{equation}

The \emph{query} set $\mathcal{Q} \subset \mathbb{D}^{train}$ has $T$ samples and can be denoted as
\begin{equation}
    \mathcal{Q} = \{(x_{_{N \times K+1}}^{image},\ x_{_{N \times K+1}}^{text},\ y_{_{N \times K+1}}),\dots,(x_{_{N \times K+ T}}^{image},\ x_{_{N \times K+T}}^{text},\ y_{_{N \times K + T}})\}
\end{equation}
Specifically, in the training stage, data labels are provided for both support set $\mathcal{S}$ and query set $\mathcal{Q}$.
Given testing data $\mathbb{D}^{test}$, our goal is to train a classifier that can map the query samples from $\mathcal{Q} \subset \mathbb{D}^{test}$   to the corresponding labels accurately with few support samples from $\mathcal{S} \subset \mathbb{D}^{test}$. 
\subsection{Main Framework}
\begin{figure}[h!]
  \includegraphics[scale=0.245]{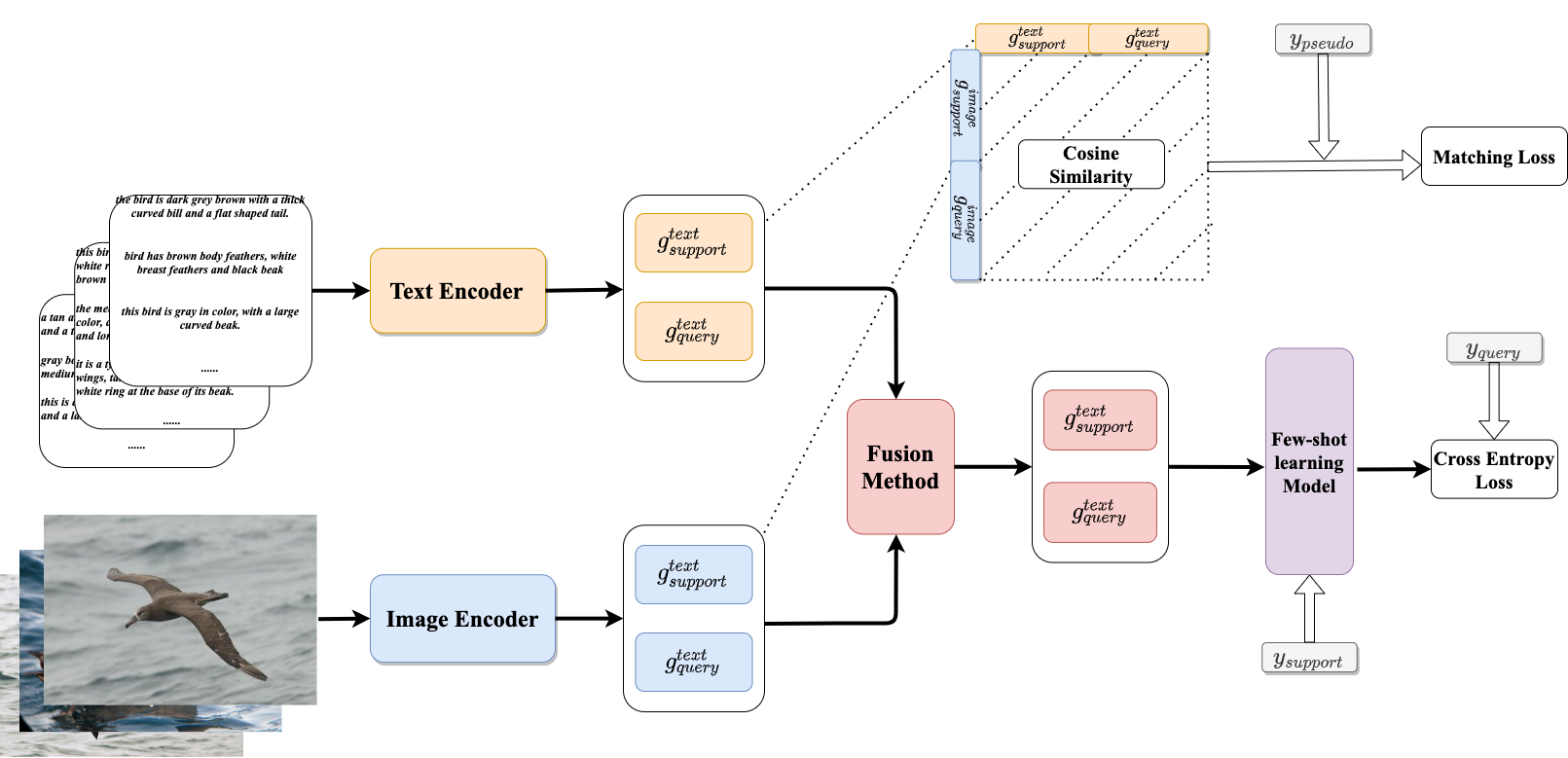}
  \caption{An overview flowchart of our main framework.}
  \label{fig:framework}
\end{figure}

Figure ~\ref{fig:framework} shows the pipeline of our framework. During training, for each episode, the images and the corresponding text descriptions are passed into the image and text encoder respectively to generate image and text feature representations.  Then, these features are passed into a fusion module to generate the fused multi-modal feature representations, which will be the input of the few-shot learning model. The final loss that is to be minimized contains two parts, the cross entropy loss and the matching loss. The cross entropy loss is used to measure how well the classifier performs on the query set. The matching loss is used to enforce the learned image and text representations to be in the same latent space, and enforce the corresponding image-text pairs to be close to each other by maximizing their cosine similarity.

\subsubsection{Multi-modal Fusion Module}

In this section, we describe the four fusing methods we have designed and implemented to fuse the image and text feature representations.

\begin{figure}[h!]
  \includegraphics[scale=0.362]{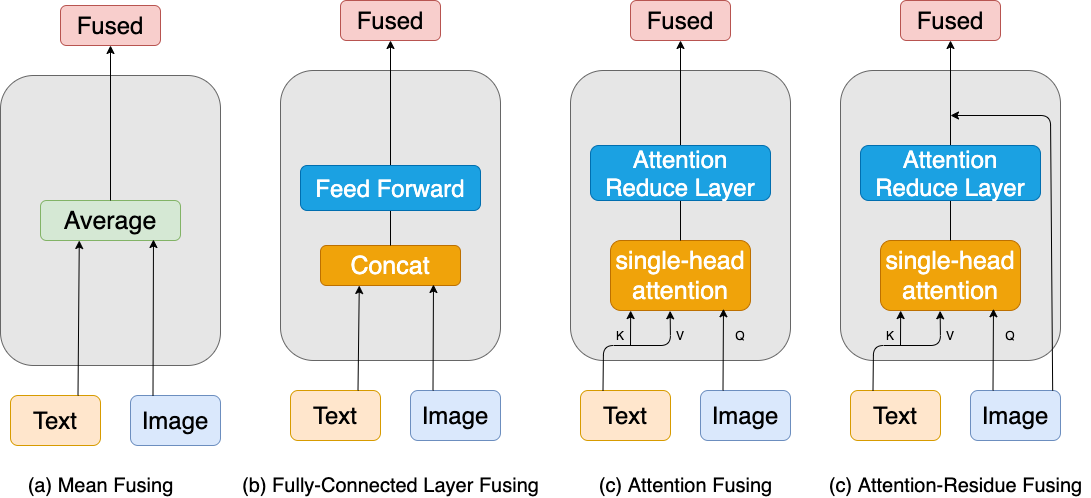}
  \caption{The four purposed multi-modal fusing methods. }
  \label{fig:fusing}
\end{figure}

\paragraph{Mean Fusion} Taking the image feature $X \in \mathbb{R}^{1 \times d}$ and text feature $Y \in \mathbb{R}^{1 \times d}$, the mean fusion simply takes the average of the two features and outputs $Z \in \mathbb{R}^{1 \times d}$. 
\paragraph{FC Fusion} Taking the image feature $X \in \mathbb{R}^{1 \times d}$ and text feature $Y \in \mathbb{R}^{1 \times d}$, the FC fusion first concatenates them into one feature vector, then passes the vector into a two-layer MLP (FC($2d$)-ReLU-Dropout(0.1)-FC($d$)) to output fused feature $Z \in \mathbb{R}^{1 \times d}$.
\paragraph{Attention Fusion} To fully utilize the power of attention mechanism in finding the correlations within a sequence of features, we modified input features into a sequence of vectors. Let's say there are $n$ sentences describing one image, instead of encoding all of the sentences into one vector as other fusing methods, we encoded each sentence separately and stacked all the features. In this case, we get the text feature $Y \in \mathbb{R}^{n \times d}$. To match the dimension of the text feature, we use $1 \times 1$ Convolution layer to reshape the image channel and get the image feature $X \in \mathbb{R}^{n \times d}$. 

The input of the single-head attention module consists of queries and keys of dimension $d_{key}$ and values of dimension $d_{value}$. For simplicity, we set $d_{key}$ and $d_{value}$ to be the same number $d$ as our embedding feature size. We first passed the text feature $Y$ into 2 separate linear layers to obtain keys $K \in \mathbb{R}^{n \times d}$ and value $V \in \mathbb{R}^{n \times d}$, and passed the image feature $X$ into another linear layer to obtain queries $Q \in \mathbb{R}^{n \times d}$. We calculated the text-guided image features through scaled dot-product attention mechanism: given a query $q \in \mathbb{R}^{1 \times d}$, key matrix $K$, and value matrix $V$, the attended feature $f \in \mathbb{R}^{1 \times d}$ is calculated by weighted summation over $V$, with weight being the attention learned between $q$ and $K$:
\begin{equation}
    f = A(q, K, V) = softmax(\frac{qK^{T}}{\sqrt{d}})V
 \end{equation}

Thus, after the single-head cross-modal attention module, we obtained the text-guided image features $Z \in \mathbb{R}^{n \times d}$. Intuitively, the attended feature $z_{i} = A(x_{i}, Y, Y)$ for $z_{i} \in Z$, $x_{i} \in X$ is obtained by reconstructing $x_{i}$ with the cross-modal similarity between $x_{i}$ to all the samples in $Y$. The last step is to reduce the attended feature $Z$ back to dimension of $\mathbb{R}^{1 \times d}$. Here we use the weighted summation over $z_{i}$ to get the final fused feature $\widetilde{Z}$:
\begin{equation}
    \widetilde{Z} = \sum^{n}_{i=1} w_{i}z_{i}
 \end{equation}
where $w=[w_{1}, w_{2},..., w_{n}] \in \mathbb{R}^{m}$ are the learned weights by passing $Z$ into a two-layer MLP (FC($d$)-ReLU-Dropout(0.1)-FC($1$)). 

\paragraph{Attention with Residual Fusion} The Attention with Residual Fusion is similar to the Attention Fusion, except that after getting the reduced attended feature $\widetilde{Z} \in \mathbb{R}^{1 \times d}$, we take the average between it and the original image feature $X \in \mathbb{R}^{1 \times d}$. Intuitively, we are mimicking the residue connection idea.

\subsection{Objective}
The loss function is defined by two types of loss. The first part is the regular classification loss for few-shot learning using Cross-Entropy. 
\begin{equation}
    \mathcal{L}_{cls} = \mathcal{L}_{CE}(\mathcal{M}(g_{support}^{fuse},\ g_{query}^{fuse},\ y_{support} | \theta_{\mathcal{M}}),\ {y_{query}})
\end{equation}
where $\mathcal{M}$ is the few-shot learning model such as ProtoNet or MAML, and $g_{support}^{fuse}$, $g_{query}^{fuse}$ are the fused multi-modal representation for samples in support set and query set. $y_{support}$, and $y_{query}$ are the ground truth labels with corresponding samples in support set and query set. $\mathcal{M}(g_{support}^{fuse},\ g_{query}^{fuse},\ y_{support} | \theta_{\mathcal{M}})$ will output the probability distribution $P( \hat{y_{query}}\ |\ g_{support}^{fuse}, \ g_{query}^{fuse}, \ y_{support})$ over classes with parameter set $\theta_{\mathcal{M}}$.\\
\newline
The second part of the loss function is the contrastive matching loss between image representation and text representation, which is inspired by \citep*{clip}. To better fuse the image representation and text representation, we would like to let the model learn the pair-wise relationship between matched images and texts, therefore we use the matching loss to constrain them.
\begin{equation}
    \mathcal{L}_{matching\_image} = \mathcal{L}_{CE}(\ matmul\ (\ g^{image},\ {g^{text}}^{T}),\ y_{pseudo})
\end{equation}
\begin{equation}
    \mathcal{L}_{matching\_text} = \mathcal{L}_{CE}(\ matmul\ (\ g^{text},\ {g^{image}}^{T}),\ y_{pseudo})
\end{equation}
\begin{equation}
    \mathcal{L}_{matching} = 0.5 * ( \mathcal{L}_{matching\_image} + \mathcal{L}_{matching\_text})
\end{equation}
\\
In this loss, $g^{image}$ and $g^{text}$ are the embeddings of image data and text data (for data in each modality, support data and query data are concatenated together), with shape $(N, d)$ for both. $N$ is the number of samples in the batch and $d$ is the embedding size. Also, paired image data and text data are aligned. $T$ is the symbol of matrix transpose and $y_{pseudo} = (1, 2, 3, \cdots , N)$ is the pseudo label with shape $(N, )$. The idea of this loss is to maximize the cosine similarity of the aligned image and text embeddings of the $N$ real pairs (the diagonal of the cosine similarity matrix) in the batch while minimizing the cosine similarity of the $N^{2} - N$ incorrect pairs in the cosine similarity matrix.
\\
\newline
Finally, we combine $\mathcal{L}_{cls}$ and $\mathcal{L}_{matching}$ together to obtain the ultimate loss.
\begin{equation}
    \mathcal{L}_{total} = \mathcal{L}_{cls} + \mathcal{L}_{matching}
\end{equation}

\subsection{Algorithm Box}
\begin{algorithm}[H]
\SetAlgoLined
\KwResult{ $\mathcal{L}_{total}$ }
 initialization: Input Batch, Feature Extractor for Image $\mathcal{F}_{image}$, Feature Extractor for Text $\mathcal{F}_{text}$, Few-shot Learning Model $\mathcal{M}$, Fusion Module $f$\\
 \While{Training}{
  Get multi-modal data $x_{support}^{image}$, $x_{query}^{image}$, $x_{support}^{text}$, $x_{query}^{text}$ and label $y_{support}$, $y_{query}$ \;
  
  Calculate multi-modal features for support data and query data: \\
  
  \ \ \ \ \ \ \ \ \ \ \ \ \ \ \ \ $g_{support}^{image}$, $g_{query}^{image}$ = $\mathcal{F}_{image} (x_{support}^{image},\ x_{query}^{image})$ \;
  
  \ \ \ \ \ \ \ \ \ \ \ \ \ \ \ \ $g_{support}^{text}$, $g_{query}^{text}$ = $\mathcal{F}_{text} (x_{support}^{text},\ x_{query}^{text})$ \;
  
  Combine multi-modal features using Fusion Module: \\
  
  \ \ \ \ \ \ \ \ \ \ \ \ \ \ \ \ $g_{support}^{fuse}$ = $f(g_{support}^{image}, \ g_{support}^{text})$ \;
  
  \ \ \ \ \ \ \ \ \ \ \ \ \ \ \ \ $g_{query}^{fuse}$ = $f(g_{query}^{image}, \ g_{query}^{text})$ \;
  
  Concatenate support data and query data for each modality and performs $l_2$ normalization \\
  
  \ \ \ \ \ \ \ \ \ \ \ \ \ \ \ \ $g^{image}$ = $Normalize(\ Concat\ (g_{support}^{image},\ g_{query}^{image}), dim=-1)$ \;
  \ \ \ \ \ \ \ \ \ \ \ \ \ \ \ \ $g^{text}$  = $Normalize(\ Concat\ (g_{support}^{text},\ g_{query}^{text}), dim=-1)$  \;
  
  Calculate the output of few-shot learning model with given inputs: \\
  
  \ \ \ \ \ \ \ \ \ \ \ \ \ \ \ \ $out$ = $\mathcal{M}(g_{support}^{fuse},\ g_{query}^{fuse},\ y_{support} | \theta_{\mathcal{M}})$ \;
  
  Calculate the classification loss from the output and query labels: \\
  \ \ \ \ \ \ \ \ \ \ \ \ \ \ \ \ $\mathcal{L}_{cls}$ = $\mathcal{L}_{CE}(out,\ {y_{query}})$ \;
  
  Calculate the similarity between image features and text features: \\
  \ \ \ \ \ \ \ \ \ \ \ \ \ \ \ \ $cos\_sim_{image}$ = $Matmul (g^{image},\ {g^{text}}^{T})$ \;

  Calculate the similarity between text features and image features: \\
  \ \ \ \ \ \ \ \ \ \ \ \ \ \ \ \ $cos\_sim_{text}$ = $Matmul (g^{text},\ {g^{image}}^{T})$ \;
  
 Calculate the contrasitive loss: \\
  \ \ \ \ \ \ \ \ \ \ \ \ \ \ \ \ $\mathcal{L}_{matching\_image}$ = $\mathcal{L}_{CE}(cos\_sim_{image},\ {y_{pseudo}})$
 
  \ \ \ \ \ \ \ \ \ \ \ \ \ \ \ \ $\mathcal{L}_{matching\_text}$ = $\mathcal{L}_{CE}(cos\_sim_{text},\ {y_{pseudo}})$ \;
   
 Combine all losses: \\
 
   \ \ \ \ \ \ \ \ \ \ \ \ \ \ \ \ $\mathcal{L}_{matching}$ = 0.5 * ($\mathcal{L}_{CE}(cos\_sim_{image},\ {y_{pseudo}})$ + $\mathcal{L}_{CE}(cos\_sim_{text},\ {y_{pseudo}})$)
   
   \ \ \ \ \ \ \ \ \ \ \ \ \ \ \ \ $\mathcal{L}_{total}$ = $\mathcal{L}_{cls}$ + $\mathcal{L}_{matching}$ \;
   
 Back Propagation to Update \;

} 
 \caption{Training Loop}
\end{algorithm}

\section{Experiments}
\subsection{Dataset Setup}
We evaluated our method on \textbf{\textit{cub-200-2011}} \citep*{cub} and \textbf{\textit{102 Category Flower Dataset}} \citep*{oxford102} datasets. Both of them  are  initially  designed  for  fine-grained  classification.  The cub-200-2011 dataset contains 11,788 images from 200 different bird species, and 102 Category Flower Dataset contains 8,189 images from 102 flower categories. Thanks to \citep*{cub_text}, 10 textual descriptions (sentences) for each image are provided in these two datasets. The descriptions are labeled use the Amazon Mechanical Turk (AMT) platform. Each description involves visual appearance for at least 10 words, and avoids naming the species and the backgrounds \citep*{cub_text_explain}. We follow the split of the datasets from \citep*{cub_text}. For cub-200-2011, we have 100 classes for meta-train, 50 classes for meta-val and 50 classes for meta-test. For 102 Category Flower Dataset, we have 82 classes for meta-train (62 classes) and meta-val (20 classes), and 20 classes for meta-test. Note that classes in meta-train, meta-val and meta-test set are disjoint.
\subsection{Experiment Setup}
\subsubsection {Feature Extractors}
\paragraph{Image Encoder} We selected two widely used backbones to extract features, ConvNet by \citep*{protonet} and ResNet12 by \citep*{metaopt}. The ConvNet backbone consists of 4 convolution blocks, and each block has the structure of Conv-BN-ReLU-Maxpool. We add a fully-connected layer in the end to obtain desired embedding size. ResNet12 is the same with the one described in \citep*{resnet} with slightly modification by \citep*{metaopt}. It has four basic residual blocks, and each of them has 3 convolution layers with Batch Normalization and ReLU non-linearity plus a skip connection.

\paragraph{Text Encoder} We selected Sentence Transformers \citep*{sentence_transformer} to extract text features from a sentence of any length. We chose the pre-trained distil-BERT \cite{sanh2020distilbert} model because of its significantly smaller parameter size, while being able to achieve  comparable good performance to those larger models. We added a fully-connected layer in the end and only fine-tuned the last layer during training.

\subsubsection{Few-shot Learning Models}
We chose two well-known few-shot learning models to run our experiments - ProtoNet \citep*{protonet} and MAML \citep*{maml} as the representatives of two coarse type of few-shot learning, metric based few-shot learning model and gradient-based few-shot learning model. We selected the euclidean distance as the metric for ProtoNet and we set the inner step size of MAML to be 0.5. The implementation detail of these two models references from Torchmeta project \citep*{torchmeta}.
\subsubsection{Training Schema}
We performed regular data augmentation with resize (to 84 x 84), center crop and normalization before training the model. The Adam optimizer is adopted in all experiments with the initial learning rate of $1\mathrm{e}{-3}$ with weight decay of $1\mathrm{e}{-3}$ as well. We decayed the learning rate by half per 80 epochs, and we trained the model for 500 epochs with a 1080TI. We set the embedding size for both image and text feature extractors to be 128.
\subsubsection{Evaluation   Protocols}
We trained our model in the meta-train dataset, and validated our model every 50 epochs in the meta-val dataset. We trained and evaluated our model in 5way-1shot setting, following  the  evaluation  process  of  previous approaches \citep*{protonet}, \citep*{maml}, \citep*{metaopt}, \citep*{mace}.  We took the model which has the best performance in the meta-val set, and randomly sampled 600 episodes from the meta-test set to test, then we reported the mean accuracy (in \% ) as well as the 95\% confidence interval.\\\\
Please note that there are various ways to disjointedly split train/val/test set of cub-200-2011 dataset in previous work. \citep*{cub_text}, \citep*{stackgan}, and \citep*{mmpfsl} follow the same fixed split (the split we used), but most of them do not use the dataset for few-shot learning. Splits from \citep*{closer}, \citep{mace} and \citep*{dpgn} use the cub-200-2011 dataset for few-shot learning, but the splits are quite random. So, it is not wise to compare their test results together. Specifically, we ensured our codebase's result of ProtoNet model under 5way-1shot setting with ConvNet backbone and image data aligning with the result of \citep*{closer}, \citep{mace}, \citep*{mmpfsl} in certain range (46\% to 50\%). Then we used our codebase's test result of ProtoNet with ConvNet backbone, MAML with ConvNet backbone and ProtoNet with ResNet12 backbone as single modality baselines, and compared them with multi-modality results with 4 different fusion methods we proposed in the previous section, as shown in Table \ref{main_table}. 
\newpage
\subsection{Experiment Results}
\paragraph{Main Results} We compared our multi-modal models with the baseline models (have a $"-"$ marker in Fusion Method Column) in the Table \ref{main_table}, which includes two backbones (ConvNet and ResNet12) and two widely used Few-shot Learning models (ProtoNet and MAML).

\begin{table}[!h]
\centering 
\caption{Few-shot classification accuracies on \textit{cub\_200\_2011}. $^{\dagger}$ denotes the result obtained using our codebase, which is comparable with the result from other papers \citep*{closer}, \citep{mace}, \citep*{mmpfsl} with same settings.}
\label{main_table}
\setlength{\tabcolsep}{0.8mm}{
\begin{tabular}{c  c  c  c  c  c}
\hline
ID & Backbone & Model & Modality & Fusion Method & Accuracy \\ \hline 
0 & ConvNet & ProtoNet & Image Only   & - &  46.99$\pm$0.77 $^{\dagger}$\\
1 & ConvNet & ProtoNet & Image + Text & Mean & 75.52$\pm$0.64 \\
2 & ConvNet & ProtoNet & Image + Text & FC & 73.41$\pm$0.66 \\
3 & ConvNet & ProtoNet & Image + Text & Attention & \textbf{78.40$\pm$0.81} \\
4 & ConvNet & ProtoNet & Image + Text & Attention with Residual & 63.60$\pm$0.73  \\ \hline 
5 & ResNet12 & ProtoNet & Image Only   & - & 53.65$\pm$0.92 \\
6 & ResNet12 & ProtoNet & Image + Text & Mean & 76.87$\pm$0.84 \\
7 & ResNet12 & ProtoNet & Image + Text & FC & 75.63$\pm$0.64 \\
8 & ResNet12 & ProtoNet & Image + Text & Attention & \textbf{77.98$\pm$0.64} \\
9 & ResNet12 & ProtoNet & Image + Text & Attention with Residual & 67.08$\pm$0.83  \\ \hline 
10 & ConvNet & MAML & Image Only   & - & 49.75$\pm$ 0.75 $^{\dagger}$\\
11 & ConvNet & MAML & Image + Text & Mean & 51.10$\pm$0.70 \\
12 & ConvNet & MAML & Image + Text & FC & \textbf{53.97$\pm$0.82} \\
13 & ConvNet & MAML & Image + Text & Attention & Fail to Converge \\
14 & ConvNet & MAML & Image + Text & Attention with Residual & Fail to Converge  \\ \hline

\end{tabular}}
\end{table}

From the experiment results, attention based fusion methods perform best, and the simplest (parameter free) fusion method - "mean" fusion works surprisingly well. In fact, fc-based fusion methods obtain a lower result comparing with "mean" fusion in metric-based few-shot learning model.  Multi-modal data fusion improves the classification accuracy of metric-based few-shot learning model a lot (vary from 24.3\% to 31.4\% ), but does not improve the results very much on the gradient-based few-shot learning methods. What's more, attention-based fusion methods fail to converge on MAML, the sensitivity of MAML model and overwhelmed parameters of attention-based fusion method could be a reason make these combinations hard to train. ResNet12 backbone performs better than ConvNet backbone consistently in most cases, the reason could be that it has better expressive power than 4-layer ConvNet since it has much more parameters, 
and these extra parameters will not cause the overfitting. 

\subsection{Additional Results}

We also conducted the experiments on another dataset to prove our findings, and the results are shown in the Table \ref{addition_table}. In this additional experiment, we chose ConvNet as our image encoding backbone and ProtoNet as the Few-shot Learning models.

\begin{table}[!h]
\centering 
\caption{Few-shot classification accuracies on \textit{oxford\_flowers102}. $^{\dagger}$ denotes the result obtained using our codebase, and it is comparable with the results from other papers \citep*{closer}, \citep{mace}, \citep*{mmpfsl} with same settings.}
\label{addition_table}
\setlength{\tabcolsep}{0.8mm}{
\begin{tabular}{c  c  c  c  c  c}
\hline
ID & Backbone & Model & Modality & Fusion Method & Accuracy \\ \hline 
0 & ConvNet & ProtoNet & Image Only   & - &  58.43$\pm$0.93 $^{\dagger}$\\
1 & ConvNet & ProtoNet & Image + Text & Mean & 75.89$\pm$0.85 \\
2 & ConvNet & ProtoNet & Image + Text & FC & 75.06$\pm$0.83 \\
3 & ConvNet & ProtoNet & Image + Text & Attention & \textbf{78.33$\pm$0.80} \\
4 & ConvNet & ProtoNet & Image + Text & Attention with Residual & 61.19$\pm$0.87  \\ \hline 

\end{tabular}}
\end{table}

We observed very similar results with the main result:  multi-modal data fusion improves the baseline performance by a lot, and attention-based fusing method performs the best. It further proves that fusing multi-modal data can be essential for improving the meta-learning performance.

\subsection{Visualization}
\begin{figure}[h!]
  \includegraphics[scale=0.35]{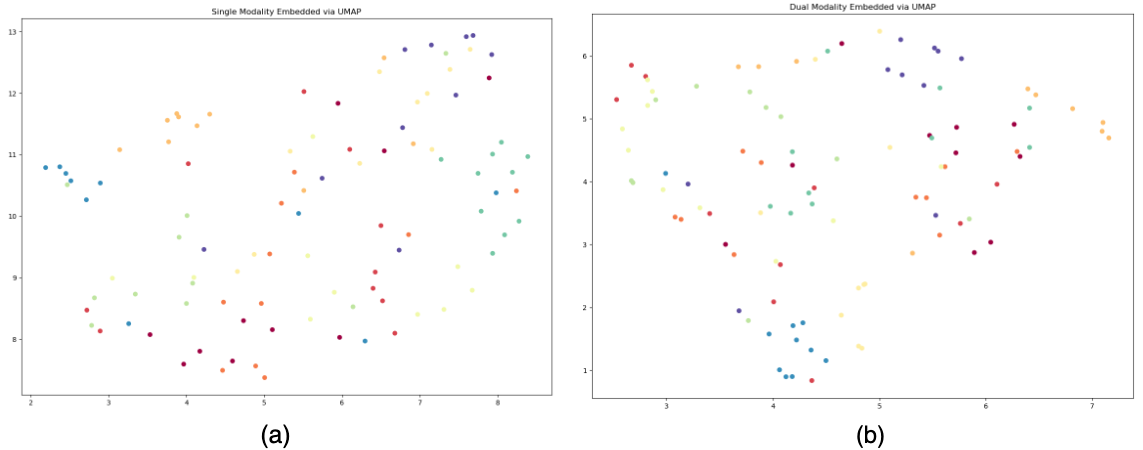}
  \caption{(a) UMAP visualization of feature space learns from single-modal data. (b) UMAP visualization of feature space learns from multi-modal data.}
  \label{fig:visual}
\end{figure}

We visualized the feature space of experiment 0 (single modality, the baseline) and experiment 3 (Multi-modality, the trail with best performance over all fusion methods) using UMAP with cosine similarity metric and neighbours of 5. The reasons we selected UMAP to visualize the feature space are UMAP could better preserve the global structure comparing to TSNE, and cost less computational resource. We randomly selected 10 classes from meta-test set, and 10 images for each class to obtain the features. Figure \ref{fig:visual} (b) shows a better separability in feature space than Figure \ref{fig:visual} (a), which means the fused feature from multi-modal data learns a more discriminate representation than single modal data.

\section{Limitation and Conclusion}
In this project, we investigated how multi-modal data improves the few-shot learning models with different fusion methods. From the results, attention-based fusion method works best (around 30 \% improvement compare with baseline), and it is hard to train a gradient-based model such as MAML if we combine the feature from another modality. We varied the image feature extractors and few-shot learning models and conducted many experiments based on them, but the study on different text feature extractor has not been widely done, and that's a limitation of this project. Another limitation is that we haven't tune the model very much due to the time and resource constraints. For future improvement, besides the investigation of different text encoders, we would like to do some research on the unified models (such as transformer) which have potential to take both image and text data into consideration.   

\newpage
\bibliographystyle{abbrv}
\bibliography{neurips_2020}

\end{document}